\documentclass[amsmath,amssymb,aps,prl,twocolumn,groupedaddress, nobibnotes,floatfix]{revtex4-2}
\usepackage{graphicx}
\usepackage{color}
\usepackage{bm}
\usepackage{amsmath}
\usepackage{slashed}
\usepackage{mathrsfs}
\usepackage[normalem]{ulem}
\usepackage{comment}
\usepackage{booktabs}

\usepackage{dcolumn}
\usepackage{hyperref}
\hypersetup{colorlinks=true, citecolor=blue, urlcolor=blue, linkcolor=blue}
\usepackage{epstopdf}

\DeclareMathOperator{\spec}{spec}
\usepackage{extarrows}

\usepackage{tikz}
\usepackage{pgfplots}
\usepgfplotslibrary{colormaps} 
\pgfplotsset{compat=1.18}
\usetikzlibrary{calc}

\begin{document}
\title{Residual spectral instabilities in representation learning}

\author{Zhen Li}
\email{li-zhen@g.ecc.u-tokyo.ac.jp}
\affiliation{Department of Complexity Science and Engineering, Graduate School of Frontier Sciences, The University of Tokyo, Kashiwa 277-8561, Japan.}
\begin{abstract}
    Learned representations can lose latent degrees of freedom successively, suggesting a cascade of transitions whose underlying stability principle remains unclear.
    Here we formulate dimension-wise posterior collapse in variational autoencoder (VAE) as a fluctuation theory around partially collapsed states.
    Interpreting the negative evidence lower bound as an effective free energy, its quadratic expansion defines a Gaussian theory whose Hessian acts as a mass matrix for latent fluctuations.
    We show that the collapsed directions form an invariant fluctuation sector and derive its exact mass spectrum in terms of a conditional residual operator.
    A local reactivation direction lowers the free energy when the decoder variance falls below the residual spectral upper edge, with equality marking marginality.
    The criterion recovers principal component thresholds in the linear Gaussian VAE limit. Viewed in reverse along continuously connected branches, the reactivation boundary provides a local criterion for successive collapse.
    Numerical continuation experiments show successive loss of latent dimensions near these spectral marginalities.
    These results support a spectral cascade interpretation governed by residual information left unexplained by the surviving representation.
\end{abstract}
\maketitle

\textit{Introduction}---
Learning a representation amounts to retaining and organizing a restricted set of degrees of freedom from a high dimensional data distribution~\cite{bengio2013representation,lecun2015deep,strouse2017deterministic,hinton2006reducing,rumelhart1986learning}.
A fundamental question is therefore how these degrees of freedom are lost as the balance between information preservation and regularization is varied~\cite{strouse2017deterministic}.
From a statistical physics perspective, this question can be approached through the stability of representations in which some latent coordinates have already collapsed~\cite{chechik2003information,wu2020phase}.
A soft mode in such a state identifies a local route by which a dormant coordinate can resume carrying information, analogous to instabilities of soft modes in structural phase transitions~\cite{cochran1960crystal,scott1974soft,casella2021soft,chechik2003information,wu2020phase}.
Continuous connections between active and collapsed branches then relate this reactivation boundary to the loss of a latent degree of freedom under reverse parameter variation.
A broader challenge in representation learning is to understand how the information retained by an existing representation shapes the spectral stability boundaries for subsequent changes in its effective dimensionality.

Variational autoencoders (VAEs) provide a natural setting in which to address this question~\cite{kingma2013auto,rezende2014stochastic}.
A VAE represents high dimensional data $\bm{x}$ through a lower-dimensional stochastic latent variable $\bm{z}$.
The encoder $q_\phi(\bm{z}|\bm{x})$ maps each input to a distribution in latent space, while the decoder $p_\theta(\bm{x}|\bm{z})$ reconstructs the data from the resulting representation.
Learning therefore balances two competing tendencies: preserving sufficient information in $\bm{z}$ to reconstruct the data and regularizing its distribution toward a prescribed prior $p(\bm{z})$.
This competition is encoded in the evidence lower bound (ELBO), which combines a reconstruction term with a Kullback--Leibler penalty that drives the variational posterior toward the latent prior~\cite{kingma2013auto,rezende2014stochastic,hoffman2016elbo,girin2022dynamical,blei2017variational,diederik2019introduction}.
A sufficiently strong relative penalty can render latent variables uninformative, producing posterior collapse~\cite{li2026posterior,lucas2019don,dai2020usual}.
This phenomenon has been attributed to optimization dynamics, expressive decoders, inference mismatch, rate, distortion, and intrinsic properties of the variational objective~\cite{he2018lagging,huang2018improving,higgins2017betavae,alemi2018fixing,lucas2019don}.
Analytical studies of linear VAEs have further revealed a close connection between collapse and the principal-component spectrum of the data~\cite{tipping1999probabilistic,lucas2019don,dang2024beyond,zihao2022advances}.
More generally, the negative ELBO can be viewed as an effective free energy functional, whose extrema define competing learned representations~\cite{carleo2019machine,iten2020discovering}.
Within this picture, the fully collapsed state is a trivial stationary solution whose local quadratic stability is controlled by the largest eigenvalue of the data covariance matrix~\cite{li2026posterior}.
This identifies a local stability boundary intrinsic to the variational landscape.

Complete collapse, however, conceals a richer structure.
In practice, latent variables need not become uninformative simultaneously: individual posterior variances can approach their prior values at distinct control parameters, leaving intermediate states in which only a subset of latent directions remains active.
Such dimension-wise collapse has been observed in both linear Gaussian VAEs and nonlinear ones, but the covariance criterion for the fully collapsed state does not determine the stability of these partially collapsed representations~\cite{lucas2019don,li2026posterior}.
This hierarchy motivates an analogy with successive symmetry breaking~\cite{chaikin1995principles,landau1936theory,rose1990statistical,chechik2003information,wu2020phase}.
Complete collapse provides a symmetric reference background, invariant under sign reversals of every latent coordinate, while partially collapsed states retain this invariance in their collapsed sectors. 
Reactivation through a posterior-mean mode can break the corresponding reflection symmetry by developing coupled encoder and decoder fields.
A sequence of such activations suggests a cascade from the fully collapsed background toward states with more active coordinates~\cite{chechik2003information,wu2020phase,rose1990statistical,rose1998deterministic}.
The reverse sequence offers a picture of dimension-wise collapse as successive symmetry restoration.
This analogy motivates our analysis of quadratic fluctuations about partially collapsed backgrounds.

The active variables already encode part of the data fluctuations. Residual information left unresolved by this representation can make the reactivation of a collapsed coordinate locally favorable.
This raises two related questions: \textit{What residual fluctuations can make a collapsed latent coordinate locally favorable to reactivate, and how do the resulting stability boundaries organize successive collapse?}

Here we construct a Gaussian fluctuation theory around partially collapsed stationary states of Gaussian VAEs.
Fluctuations associated with collapsed latent directions form an invariant sector of the Hessian, whose stability is governed by a conditional residual operator.
We determine when the quadratic form in the collapsed sector acquires a negative direction. Its marginal boundary identifies a possible continuous bifurcation between stationary backgrounds with different numbers of active latent fields.
This criterion recovers the covariance threshold for complete collapse and the principal-component thresholds of linear Gaussian VAEs.
This local fluctuation analysis suggests a cascade of spectral marginalities, whose relation to successive collapse is examined numerically in nonlinear VAEs.

\textit{Gaussian fluctuations}---
We consider a Gaussian VAE~\cite{rezende2018taming} with data $\bm x \in \mathbb{R}^N$ and latent variables $\bm z \in \mathbb{R}^n$, $n<N$.
The latent prior, decoder, and diagonal variational posterior are
\begin{align}
    p(\bm z) &= \prod_{j=1}^{n}\mathcal{N}(z_j;0,1),\\
    p_{\theta}(\bm x | \bm z) &= \prod_{i=1}^{N}\mathcal{N}(x_i;f_i(\bm z),\sigma'^2),\\q_{\phi}(\bm z | \bm x) &= \prod_{j=1}^{n}\mathcal{N}(z_j;\mu_j(\bm x),\sigma_j^2(\bm x)),
\end{align}
respectively.
Here $\sigma'^2$ is the decoder variance and serves as the control parameter.
Up to an additive constant, we write the negative evidence lower bound as an effective free energy functional~\cite{dai2020usual,ichikawa2025high},
\begin{align}
\mathcal F[\Phi]=&\frac{1}{2\sigma'^2}\left< \int d \bm z\, q_\phi(\bm z|\bm x) \|\bm x-\bm f(\bm z)\|^2\right>_{\bm x} \\
&+ \frac{1}{2} \left<\sum_{j=1}^{n}\left[\mu_j^2(\bm x)+\sigma_j^2(\bm x)-\ln\sigma_j^2(\bm x)\right]\right>_{\bm x},
\end{align}
where $\Phi := \{\bm f, \bm\mu, \bm\sigma\}$ denotes the learned fields.
Increasing $\sigma'^2$ weakens the reconstruction term relative to the latent regularization and therefore favors collapsed representations.

The conventional fully collapsed state corresponds to $f_i = \left<x_i\right>_{\bm x}$, $\mu_j = 0$ and $\sigma_j = 1$ for all $i$ and $j$~\cite{li2026posterior}.
Here we instead consider a generic stationary state in which the latent space is partitioned into an active set $A$ and a collapsed set $C$.
Such a state, consistent with the stationary equation~\cite{li2026posterior}, satisfies
\begin{eqnarray}
    \bar{\bm f}(\bm z) = \bar{\bm f}^A(\bm z_A),
\end{eqnarray}
and for $c \in C$,
\begin{eqnarray}
    \bar{\mu}_c(\bm x)=0,\quad \bar{\sigma}_c(\bm x)=1,
\end{eqnarray}
so that
\begin{eqnarray}
    \bar q(\bm z|\bm x)= \bar q_A(\bm z_A|\bm x)\varphi_C(\bm z_C),
\end{eqnarray}
with $\varphi_C$ the standard Gaussian measure in the collapsed subspace.
The active variables therefore retain a nontrivial representation of the data, whereas the collapsed variables carry no information about $\bm x$.

To analyze local stability around this partially collapsed stationary state, we expand the effective free energy around $\bar{\Phi}$:
\begin{eqnarray}
    \mathcal F[\bar\Phi+\delta\Phi]=\mathcal F[\bar\Phi]+\frac{1}{2}\left<\delta\Phi,\bm K\,\delta\Phi\right>+O(\delta\Phi^3),\label{eq:second_var}
\end{eqnarray}
where $\bm K := \delta^2\mathcal{F} / \delta\Phi^2 |_{\bar\Phi}$ is the Hessian.
Equation~\eqref{eq:second_var} defines a Gaussian fluctuation theory around the learned state, with the eigenvalues of $\bm K$ playing the role of squared masses~\cite{mussardo2010statistical}.
A zero eigenvalue identifies a marginal fluctuation mode, whereas a negative eigenvalue indicates a direction of decreasing free energy~\cite{cochran1960crystal,bruce1981structural}.

The factorization in Eq.~\eqref{eq:second_var} imposes a nontrivial structure on these fluctuations.
We decompose a decoder fluctuation into the component that is independent of the collapsed coordinates and its orthogonal complement,
\begin{align}
    \delta\bm f &= P_A \delta\bm f + (1-P_A) \delta\bm f \equiv \delta\bm f^{\parallel} + \delta\bm f^{\perp},
\end{align}
where $P_A$ denotes averaging over the Gaussian measure $\varphi_C$.
Together with the encoder fluctuations, this defines
\begin{eqnarray}
    \delta\Phi_A&:=(\delta\bm f^\parallel,\delta\bm\mu_A,\delta\bm\sigma_A),\\ \delta\Phi_C&:=(\delta\bm f^\perp,\delta\bm\mu_C,\delta\bm\sigma_C),
\end{eqnarray}
and remarkably, the quadratic theory separates exactly,
\begin{eqnarray}
    \bm K = \bm K_A \oplus \bm K_C.
\end{eqnarray}
Thus, fluctuations that activate a collapsed latent direction form an \textit{invariant sector} of the full Hessian.
Local stability against reactivation of collapsed directions can therefore be analyzed independently of the fluctuations of active sectors.
It's worth noting that the stability of the full stationary state additionally requires $\bm K_A\succeq0$.

\textit{Residual spectrum}---
To determine the stability of the collapsed sector, we characterize the data fluctuations left unresolved by the active representation.

At a stationary point, variation with respect to the decoder gives the conditional mean~\cite{li2026posterior}, $ \bar{\bm f}^A(\bm z_A)=\mathbb{E}[\bm x|\bm z_A] $, where the expectation is taken with respect to the joint distribution $p_{\rm data}(\bm x)\bar{q}_A(\bm z_A|\bm x)$.
The corresponding residual field,
\begin{eqnarray}
    r_i(\bm x,\bm{z}_A):=x_i-\bar{f}_i^A(\bm{z}_A),
\end{eqnarray}
therefore represents the part of the data fluctuations left unresolved by the active latent representation.
To characterize which of these residual fluctuations can be captured by an additional latent degree of freedom, we introduce the conditional residual operator
\begin{eqnarray}
    (\mathcal R_A h)_i(\bm z_A)=\mathbb E\left[r_i(\bm x,\bm{z}_A)h(\bm x)\mid\bm{z}_A\right],
\end{eqnarray}
which maps a fluctuation $h(\bm x)$ in data space to its residual correlation at fixed $\bm{z}_A$.
Its positive semi-definite Gram operator,
\begin{eqnarray}
    \mathcal{C}_A := \mathcal{R}_A^\dagger\mathcal{R}_A,\quad \Lambda_*^{(A)}:=\sup\spec\mathcal{C}_A=\|\mathcal{R}_A\|^2,
\end{eqnarray}
defines the residual spectrum of the partially collapsed representation.

The spectral upper edge has a useful variational interpretation~\cite{scholkopf1998nonlinear}. Let $\rho_A$ denote the active latent marginal induced by $p_{\rm data}\bar q_A$. Since the stationary decoder is the conditional mean, $\mathcal R_Ah=\operatorname{Cov}(\bm x,h(\bm x)\mid\bm z_A)$, and hence
\begin{equation}
    \Lambda_*^{(A)}=
    \sup_{\langle h^2\rangle_{\bm x}=1}
    \mathbb E_{\rho_A}\!\left[
    \left\|\operatorname{Cov}(\bm x,h(\bm x)\mid\bm z_A)\right\|^2
    \right].\label{eq:residual-variational}
\end{equation}
Thus the softest direction is selected by the strongest conditional correlation between a normalized data fluctuation and the unexplained residual.
This optimization extends the principal components to nonlinear data fluctuations and their conditional residual correlations at each active latent state.
The resulting spectrum characterizes residual structure conditioned on the surviving representation.
Its dependence on the learned joint distribution is essential: data covariance alone does not generally specify these conditional correlations.

The physical significance of $\mathcal{C}_A$ follows from the Gaussian fluctuation theory developed above.
Expanding decoder fluctuations in Hermite modes of a collapsed latent coordinate, the first Hermite mode couples to fluctuations of its posterior mean, while the second couples to fluctuations of its posterior variance.
All remaining modes are strictly massive.
For each residual eigenmode $\mathcal{C}_{A} \psi_\alpha = \Lambda_\alpha^{(A)}\psi_\alpha$, the coupled fluctuations reduce to independent mass blocks.
For example, normalize a positive-eigenvalue mode by $\langle\psi_\alpha^2\rangle_{\bm x}=1$. Its decoder partner is $\bm\chi_\alpha=\mathcal R_A\psi_\alpha/\sqrt{\Lambda_\alpha^{(A)}}$, with $\mathbb E_{\rho_A}\|\bm\chi_\alpha\|^2=1$.
For a collapsed coordinate $c$, the perturbations $\delta\bm f^\perp=u z_c\bm\chi_\alpha$ and $\delta\mu_c=v\psi_\alpha$ contribute $\tfrac12(u,v)\bm M_{\mu,\alpha}(u,v)^{\mathsf T}$ to the free-energy expansion, where
\begin{equation}
    \bm M_{\mu,\alpha}=
    \begin{pmatrix}
        1/\sigma'^2&-\sqrt{\Lambda_\alpha^{(A)}}/\sigma'^2\\
        -\sqrt{\Lambda_\alpha^{(A)}}/\sigma'^2&1
    \end{pmatrix}.\label{eq:mean-mass-block}
\end{equation}
The diagonal terms penalize decoder and encoder distortions separately; residual correlations couple them and can make a coordinated fluctuation favorable. 
In this normalized field metric, the lower squared mass is
\begin{equation}
    m^2_{\mu,\alpha,-}=
    \frac{1+\sigma'^2-\sqrt{(1-\sigma'^2)^2+4\Lambda_\alpha^{(A)}}}{2\sigma'^2}.
    \label{eq:mean-mass-lower}
\end{equation}
Since $\det\bm M_{\mu,\alpha}=(\sigma'^2-\Lambda_\alpha^{(A)})/\sigma'^4$ and its trace is positive, the lower mode changes sign precisely at the residual spectral threshold.
The posterior-width channel couples to the second Hermite mode and has the same marginality condition; its mass block and the remaining spectrum are given in the Supplemental Material.
Their lowest eigenvalues soften and vanish precisely at $\sigma'^2 = \Lambda_\alpha^{(A)}$.
Thus, for a nonempty collapsed set $C$ and a bounded residual operator, nonnegative curvature in the collapsed sector is equivalent to
\begin{eqnarray}
    \bm{K}_C \succeq 0 \quad \Longleftrightarrow \quad \sigma'^2 \geq \Lambda_*^{(A)}.\label{eq:k_c_stability}
\end{eqnarray}

Equivalently, Eq.~\eqref{eq:k_c_stability} locates the onset of negative curvature in the collapsed fluctuation sector:
\begin{equation}
 \exists\,\delta\Phi_C\ s.t.\ 
 \langle\delta\Phi_C,\bm K_C\delta\Phi_C\rangle<0\,\Longleftrightarrow\,\sigma'^2<\Lambda_*^{(A)}.\label{eq:reactivation}
\end{equation}
The unstable fluctuation is a coupled deformation of the encoder and decoder fields, with their mixing set by residual correlations. The quadratic operator contains an identical set of mass blocks for each collapsed coordinate, so a negative direction can be supported in a single coordinate sector. This establishes a local instability toward reactivation of at least one collapsed field. Higher-order terms govern the continuation to finite amplitude, while the evolution dynamics determine which stationary branch is approached.

For $\sigma'^2>\Lambda_*^{(A)}$, this sector has strictly positive curvature; equality closes its spectral gap and yields a zero-mass mode when the spectral supremum is attained by an eigenmode.

The familiar criterion for complete posterior collapse follows as a limiting case~\cite{li2026posterior}.
For $A = \varnothing$, no active representation survives, $\bar{\bm{f}}^\varnothing$ reduces to the data mean, and the nonzero residual spectrum coincides with the covariance spectrum $\bm\Sigma_{X}$ of the data~\cite{blanchard2007statistical}.
Hence,
\begin{eqnarray}
    \Lambda_*^{(\varnothing)} = \lambda_{\rm max}\left(\bm\Sigma_{X}\right),
\end{eqnarray}
recovering the full-collapse threshold obtained previously~\cite{li2026posterior}.
The covariance criterion is therefore the terminal member of a more general hierarchy: before complete collapse, stability is controlled by the spectrum of fluctuations left unexplained by the current representation rather than by the spectrum of the original data alone.

To relate the reactivation criterion to collapse, consider a coordinate $c$ that is active in $A$ and collapsed in $A'=A\setminus\{c\}$.
The relevant quadratic expansion is taken about the stationary background with active set $A'$.
Its modes of the collapsed sector describe coupled encoder--decoder fluctuations that reactivate $c$.
For a background with nonnegative curvature for the collapsed sector, a zero-mass mode must lie at the residual spectral upper edge [Eqs.~\eqref{eq:mean-mass-lower} and \eqref{eq:k_c_stability}]:
\begin{equation}
 \sigma'^2=\Lambda_*^{(A')}(\sigma'^2),\qquad A\to A'.
 \label{eq:cascade-connection}
\end{equation}
Interpreting an observed collapse through this condition assumes that the active branch reaches such a marginal collapsed background continuously. Under this assumption, successive losses of active coordinates can be associated with successive marginalities, each evaluated on the corresponding collapsed branch. This provides a possible local explanation of the cascade, whose consistency with numerical continuation we examine below.

\textit{Spectral cascade}---
We examine whether increasing the control parameter $\sigma'^2$ produces successive collapse near the marginal points of the corresponding collapsed backgrounds.
We use a set of 8 dimension synthetic Gaussian mixture data (see Supplemental Material) with principal components $\lambda_i=\exp[-0.1(i-1)]$ ($i=1,\ldots,8$), 
and employ a four-dimensional latent space. The encoder consists of two hidden layers of widths 256 and 128, while the decoder uses the reversed architecture, with ReLU activations throughout.
The decoder variance $\sigma'^2$ is varied as the control parameter.

\begin{figure}
    \includegraphics[width=\linewidth]{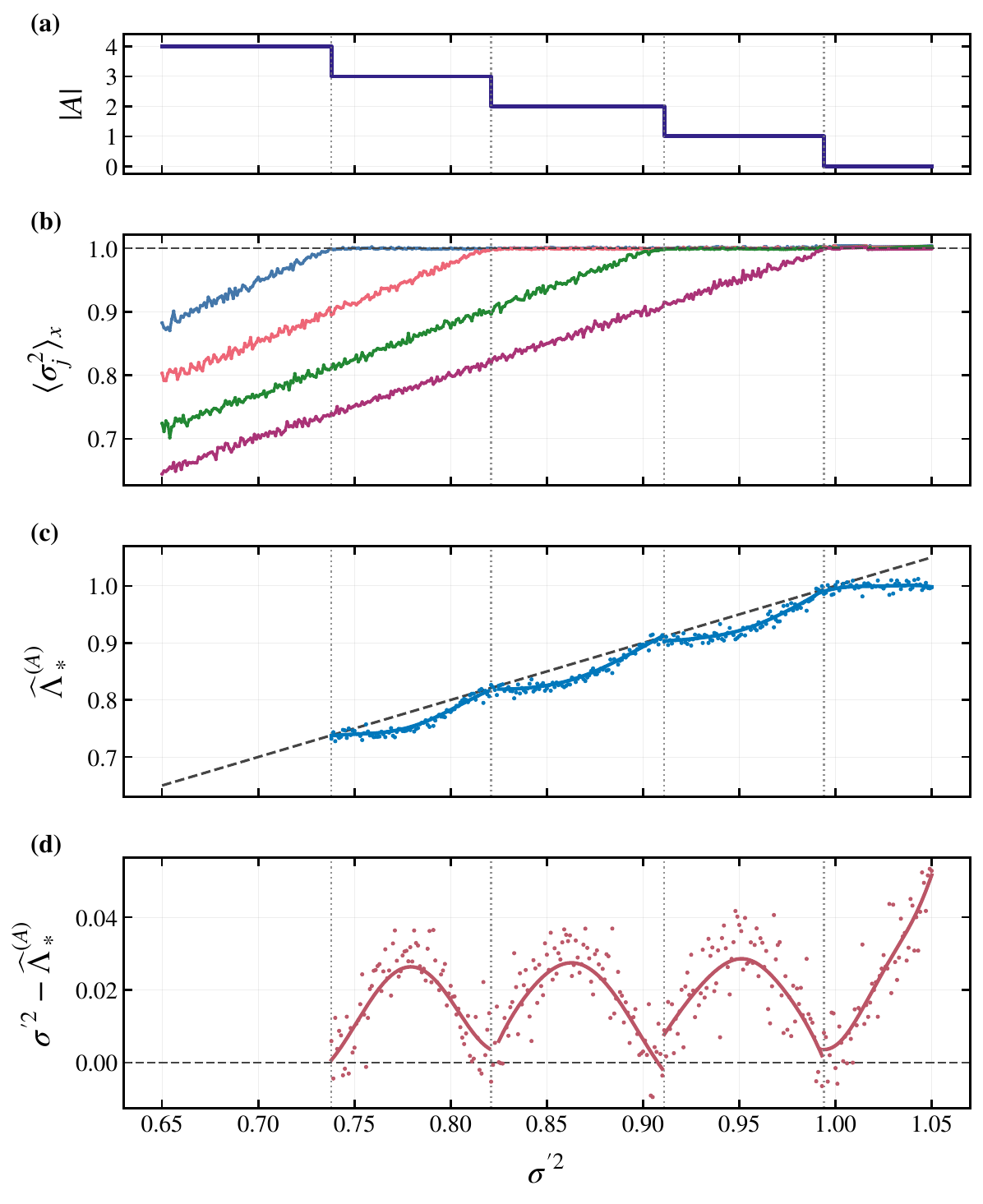}
    \caption
    {
        Spectral cascade of dimension-wise posterior collapse. 
        (a) Number of active latent dimensions $|A|$ and (b) mean posterior variances $\left<\sigma_j^2\right>_{\bm x}$ versus decoder variance $\sigma'^2$, showing showing successive loss of individual latent dimensions. 
        (c) Estimated leading residual eigenvalue $\hat{\Lambda}_*^{(A)}$, and the dashed line denotes the marginality condition $\sigma'^2 = \hat{\Lambda}_*^{(A)}$.
        (d) Estimated spectral stability margin $\sigma'^2-\hat\Lambda_*^{(A)}$, which controls the sign of the lowest squared masses of the collapsed sector.
    }
    \label{fig:mlp}
\end{figure}

Figure~\ref{fig:mlp}(a) shows the resulting effective number of active latent dimensions.
Starting from $|A|=4$ at small $\sigma'^2$, the representation loses one latent degree of freedom at a time and passes through a sequence from 4 to 0.
The individual posterior variances in Fig.~\ref{fig:mlp}(b) reveal the microscopic origin of this staircase.
Each $\left<\sigma^2_j\right>_{\bm x}$ approaches its prior value of unity at a distinct value of $\sigma'^2$, after which the corresponding latent coordinate remains collapsed.
The steps in $|A|$ therefore represent genuine dimension-wise posterior collapse rather than a smooth reduction of information distributed uniformly among all latent variables.

To probe the collapsed-sector curvature in Eq.~\eqref{eq:reactivation}, we estimate the residual spectral upper edge $\hat{\Lambda}_*^{(A)}$ along the numerical branches with a nonempty collapsed sector.
At each frozen checkpoint, we average the decoder over collapsed prior coordinates and approximate the conditional residual operator using posterior weights normalized over sampled data and Monte Carlo samples of the active latent marginal.
Power iteration on its empirical Gram operator, restricted to centered data-space fluctuations, yields the leading eigenvalue.
To reduce finite-sampling bias, repeated estimates with $B$ data points and $L$ active-latent samples are fitted to $\hat{\Lambda}(B,L)=\Lambda_\infty+b/B+c/L$ by inverse-variance weighted least squares.
The intercept defines $\hat{\Lambda}_*^{(A)}$.
Detail algorithm can be referred in the Supplemental Material.
Figure~\ref{fig:mlp}(c) compares this quantity with the spectral marginality condition $\sigma'^2 = \hat{\Lambda}_*^{(A)}$.
On intervals with fixed $A$, the estimated spectral upper edge generally lies below $\sigma'^2$, consistent with positive squared masses in the collapsed sector.
The approach to equality near the observed loss of a latent degree of freedom is consistent with a marginal connection between stationary branches.
Once the transition occurs, the active representation changes and $\hat{\Lambda}_*^{(A)}$ shifts onto a new branch.

The quantity $\sigma'^2-\hat\Lambda_*^{(A)}$ in Fig.~\ref{fig:mlp}(d) controls the sign of the lowest squared masses of the collapsed sector.
Its approach to zero signals softening about the corresponding background and is consistent with a continuous connection to a branch carrying one additional active field.
Each segment is evaluated with the residual operator of its own active background. The proximity of the observed collapse events to these marginal points supports the proposed spectral interpretation of the numerical branches.
As a direct curvature check, we perturb representative projected backgrounds along the predicted coupled encoder--decoder modes and evaluate the resulting free-energy differences. Fixed-background scans exhibit a change from negative to positive curvature near the residual spectral threshold, supporting the local fluctuation picture (see Supplemental Material).

\textit{Discussion and conclusions}---
In the linear Gaussian VAE limit, the residual spectrum can be evaluated explicitly. For an active set $A$ aligned with the principal directions of the data, the condition for local stability in the collapsed sector under linear perturbations reduces to (see Supplemental Material):
\begin{eqnarray}
    \sigma'^2 \geq \max_{\alpha\notin A}\lambda_{\alpha}(\bm\Sigma_X),
\end{eqnarray}
recovering the exact principal component thresholds known for linear VAEs~\cite{lucas2019don,dang2024beyond, tipping1999probabilistic}.
The residual spectrum therefore generalizes the fixed covariance spectrum of linear models to a representation dependent spectrum for nonlinear learned states.

Our analysis is restricted to the quadratic expansion of the effective free energy and thus describes a free, Gaussian theory of fluctuations.
The direct result is a criterion for the onset of negative curvature in the collapsed sector about a specified stationary background. Under a continuous branch connection, the reverse passage through marginality describes the disappearance of an active latent field. The sequence of realized backgrounds, the number of fields involved in each transition, and discontinuous branch changes depend on the global free-energy landscape, nonlinear fluctuation terms, and evolution dynamics.
Higher-order terms couple the fluctuation modes and govern the free-energy variation along directions whose quadratic restoring term vanishes at marginality.
Extending the expansion along these soft directions would help determine the finite-amplitude branches and the character of the associated bifurcations~\cite{wilson1974renormalization,zinn2021quantum}.

A second limitation lies in the background state itself.
Although the stability criterion has a simple spectral form, the residual operator depends on the stationary $\bar{\bm f}^A$ and $\bar{q}_A$, which obey nonlinear self-consistency equations involving high dimensional integrals. 
Except in special limits such as the linear Gaussian model, these equations cannot generally be eliminated to express the residual spectrum solely in terms of simple data statistics. 
Controlled approximations to the background state, and relations between residual spectra across successive stationary branches, therefore remain important open problems.

From the perspective of representation learning, the residual spectrum identifies fluctuations left unresolved by the active latent fields that can destabilize a collapsed sector.
It thereby connects the local availability of an additional representational degree of freedom to the information already retained by the learned background. 
Relating these spectra across stationary branches offers a route toward understanding how the effective dimension of a representation changes as the balance between reconstruction and regularization is varied. 
Extending this stability analysis to other latent-variable objectives is a natural next step toward a spectral account of the organization of learned representations~\cite{strouse2017deterministic,achille2018emergence,vincent2010stacked}.

\textit{Acknowledgement}---This work was supported by JST SPRING, Grant No. JPMJSP2108.

\bibliography{ref}

\clearpage
\onecolumngrid
\setcounter{equation}{0}
\setcounter{figure}{0}
\setcounter{table}{0}
\setcounter{section}{0}
\setcounter{subsection}{0}
\renewcommand{\theequation}{S\arabic{equation}}
\renewcommand{\thefigure}{S\arabic{figure}}
\renewcommand{\thetable}{S\arabic{table}}
\renewcommand{\theHequation}{supp.\arabic{equation}}
\renewcommand{\theHfigure}{supp.\arabic{figure}}
\renewcommand{\theHtable}{supp.\arabic{table}}
\renewcommand{\theHsection}{supp.\arabic{section}}

\begin{center}
    \bfseries
    Supplemental material for\\
    ``Residual spectral instabilities in representation learning''
\end{center}

\subsection{Objective, notation, and stationary equations}

We use the Gaussian VAE and the effective free energy $\mathcal F$
(negative ELBO) introduced in the main text. The data and latent variables are
$\bm x\in\mathbb R^N$ and $\bm z\in\mathbb R^n$, respectively, where $N>n$.
The data distribution is $p_{\rm data}(\bm x)$, and the fixed decoder
variance $\sigma'^2>0$ serves as the control parameter.
Throughout, $\sigma_j$ denotes an encoder standard deviation and
$\sigma_j^2$ its variance. We use $\lambda_\alpha$ for data covariance
eigenvalues and $\Lambda_\alpha^{(A)}$ for residual eigenvalues;
$\Lambda_*^{(A)}$ denotes the upper edge of the residual spectrum.
The latent prior, decoder, and diagonal variational posterior are
\begin{align}
 p(\bm z)&:=\prod_{j=1}^n\mathcal N(z_j;0,1),\\
 p_\theta(\bm x|\bm z)&:=\prod_{i=1}^N\mathcal N(x_i;f_i(\bm z),\sigma'^2),\\
 q_\phi(\bm z|\bm x)&:=\prod_{j=1}^n
 \mathcal N(z_j;\mu_j(\bm x),\sigma_j^2(\bm x)).
\end{align}
Up to an additive constant, the effective free energy reads
\begin{align}
 \mathcal F[\Phi]
 ={}&\frac{1}{2\sigma'^2}\int d\bm x\,d\bm z\,
 p_{\rm data}(\bm x)q_\phi(\bm z|\bm x)
       \sum_{i=1}^N[x_i-f_i(\bm z)]^2\notag\\
 &+\frac12\int d\bm x\,p_{\rm data}(\bm x)\sum_j
       [\mu_j(\bm x)^2+\sigma_j(\bm x)^2-\ln\sigma_j(\bm x)^2].
 \label{eq:free-energy}
\end{align}
We vary the learned fields $\Phi:=\{\bm f,\bm\mu,\bm\sigma\}$ independently.
Data averages are denoted by
\[
 \langle g\rangle_{\bm x}:=\int d\bm x\,p_{\rm data}(\bm x)g(\bm x).
\]

Varying the decoder mean and the Gaussian encoder parameters directly
in Eq.~\eqref{eq:free-energy} gives the stationary equations~\cite{li2026posterior}:
\begin{align}
 0&=\int d\bm x\,p_{\rm data}(\bm x)q_\phi(\bm z|\bm x)
                       [x_i-f_i(\bm z)],
 \label{eq:decoder-stationary}\\
 \mu_j(\bm x)&=-\frac{1}{2\sigma'^2}\int d\bm z\,q_\phi(\bm z|\bm x)
                  \sum_i[x_i-f_i(\bm z)]^2
                  \frac{z_j-\mu_j(\bm x)}{\sigma_j(\bm x)^2},\\
 \sigma_j(\bm x)-\sigma_j(\bm x)^{-1}
 &=-\frac{1}{2\sigma'^2}\int d\bm z\,q_\phi(\bm z|\bm x)
                  \sum_i[x_i-f_i(\bm z)]^2
                  \left[\frac{[z_j-\mu_j(\bm x)]^2}{\sigma_j(\bm x)^3}
                        -\frac{1}{\sigma_j(\bm x)}\right].
 \label{eq:encoder-stationary}
\end{align}
Specifically,
\begin{equation}
 f_i(\bm z)=
 \frac{\int d\bm x\,p_{\rm data}(\bm x)q_\phi(\bm z|\bm x)x_i}
      {\int d\bm x\,p_{\rm data}(\bm x)q_\phi(\bm z|\bm x)},
\end{equation}
is the conditional mean of $x_i$ under the joint distribution $p_{\rm data}(\bm x)q_\phi(\bm z|\bm x)$.

\subsection{Second variation at a general stationary point}

Choose a stationary solution $\bar\Phi=(\bar{\bm f}, \bar{\bm\mu}, \bar{\bm\sigma})$ as the background for a fluctuation
expansion, and let
$\bar q$ be the variational posterior evaluated on this background:
\begin{equation}
 \bar q(\bm z|\bm x)=\prod_{j=1}^n
 \mathcal N(z_j;\bar\mu_j(\bm x),\bar\sigma_j(\bm x)^2),
\end{equation}
We write the residual field, standardized latent
variable, and marginal density on the background as
\begin{align}
 r_i(\bm x,\bm z)&=x_i-\bar f_i(\bm z),\\
 \epsilon_j(\bm x,\bm z)&=
       \frac{z_j-\bar\mu_j(\bm x)}{\bar\sigma_j(\bm x)},\\
 \rho(\bm z)&=\int d\bm x\,p_{\rm data}(\bm x)\bar q(\bm z|\bm x),
\end{align}
respectively.

Denote the fluctuations by $\delta f_i(\bm z)$,
$\delta\mu_j(\bm x)$, and $\delta\sigma_j(\bm x)$, and their
collection by $\delta\Phi:=(\delta\bm f(\bm z),\delta\bm\mu(\bm x),\delta\bm\sigma(\bm x))$. Following the main text, we define the
Hessian, or mass operator, of the effective free energy by
\begin{equation}
 \bm K:=\left.\frac{\delta^2\mathcal F}{\delta\Phi^2}\right|_{\bar\Phi}.
\end{equation}
Nonnegative free-energy curvature corresponds to $\bm K\succeq0$.
The Gaussian fluctuation theory is defined by
\begin{align}
 \mathcal F[\bar\Phi+\delta\Phi]
 =\mathcal F[\bar\Phi]+\frac{1}{2}\langle\delta\Phi,\bm K\delta\Phi\rangle
 +O(\delta\Phi^3),\label{eq:sign}
\end{align}
We denote the quadratic form by $Q[\delta\Phi]:=\langle\delta\Phi,\bm K\delta\Phi\rangle$.
Here the field inner product uses $L^2(p_{\rm data}(\bm x)\bar{q}(\bm z | \bm x);\mathbb R^{N+2n})$.
To display the second variation explicitly, introduce the local field
vector $\delta\bm v(\bm x,\bm z):=
(\delta\bm f(\bm z),\delta\bm\mu(\bm x),\delta\bm\sigma(\bm x))^{\mathsf T}$.
Then
\begin{align}
 Q[\delta\Phi]
 &=\int d\bm x\,d\bm z\,p_{\rm data}(\bm x)\bar q(\bm z|\bm x)
 \delta\bm v^{\mathsf T}\bm k(\bm x,\bm z)\delta\bm v,\\
 \bm k&:=\begin{pmatrix}
        \bm k_{ff} & \bm k_{f\mu} & \bm k_{f\sigma}\\
        \bm k_{f\mu}^{\mathsf T} & \bm k_{\mu\mu} & \bm k_{\mu\sigma}\\
        \bm k_{f\sigma}^{\mathsf T} & \bm k_{\mu\sigma}^{\mathsf T} & \bm k_{\sigma\sigma}
    \end{pmatrix}.\label{eq:hessian_general}
\end{align}
The local coefficients $\bm k$ represent the free-energy quadratic form
under this joint integral; the induced operator on the weighted field
spaces is $\bm K$. The coefficients are
\begin{align}
 k_{ff}^{ii'}
 &=\frac{1}{\sigma'^2}\delta_{ii'},\\
 k_{f\mu}^{ij}
 &=-\frac{1}{\sigma'^2}
                  r_i(\bm x,\bm z)\frac{\epsilon_j}{\bar\sigma_j},\\
 k_{f\sigma}^{ij}
 &=-\frac{1}{\sigma'^2}
                  r_i(\bm x,\bm z)\frac{\epsilon_j^2-1}{\bar\sigma_j},\\
 k_{\mu\mu}^{jk}
 &=
 \delta_{jk}+\frac{1}{2\sigma'^2}\sum_i r_i^2
       \frac{\epsilon_j\epsilon_k-\delta_{jk}}{\bar\sigma_j\bar\sigma_k},\\
 k_{\mu\sigma}^{jk}
 &=\frac{1}{2\sigma'^2}\sum_i r_i^2\frac{\epsilon_j(\epsilon_k^2-1)-2\delta_{jk}\epsilon_j}
{\bar\sigma_j\bar\sigma_k},\\
 k_{\sigma\sigma}^{jk}
 &=
 (1+\bar\sigma_j^{-2})\delta_{jk}+\frac{1}{2\sigma'^2}\sum_i r_i^2
 \frac{(\epsilon_j^2-1)(\epsilon_k^2-1)+\delta_{jk}(1-3\epsilon_j^2)}
      {\bar\sigma_j\bar\sigma_k}.
 \label{eq:kernels}
\end{align}

\subsection{Partial collapse and the invariant Hessian blocks}

Besides the fully collapsed trivial solution, the stationary equations~\eqref{eq:decoder-stationary}--\eqref{eq:encoder-stationary} also give partially collapsed background,
\begin{align}
 \bar f_i(\bm z)&=\bar f_i^A(\bm z_A),\\
 \bar\mu_c(\bm x)&=0\quad(c\in C),\\
 \bar\sigma_c(\bm x)&=1\quad(c\in C),
\end{align}
where the active set $A$ and the collapsed set $C$ partition the latent indices $\{1,\ldots,n\}$.
\begin{eqnarray}
    \bm z = (\bm z_A, \bm z_C).
\end{eqnarray}
We use $a\in A$ and $c\in C$, while $i,i'$ label data components
and $j,k$ label arbitrary latent components.

Moreover, the encoder can be decomposed as
\begin{eqnarray}
    \bar q(\bm z|\bm x)=\bar q_A(\bm z_A|\bm x)\varphi_C(\bm z_C),
\end{eqnarray} 
where $\varphi_C$ is the standard $\left|C\right|$-dimensional Gaussian density, and the residual reduces to the main-text field
$r_i(\bm x,\bm z_A):=x_i-\bar f_i^A(\bm z_A)$.
We may also write the marginal density for the active latent variables on the background as
\begin{equation}
 \rho_A(\bm z_A)=\int d\bm x\,p_{\rm data}(\bm x)
                         \bar q_A(\bm z_A|\bm x).
\end{equation}

It is easy to discover that there is no active-collapse coupling in the Hessian~\eqref{eq:hessian_general} for perturbations on encoders $\delta\bm\mu$ and $\delta\bm\sigma$, by splitting them into active set $A$ and collapsed set $C$:
\begin{eqnarray}
    \delta\bm\mu &= (\delta\bm\mu_A,\delta\bm\mu_C),\\
    \delta\bm\sigma &= (\delta\bm\sigma_A,\delta\bm\sigma_C).
\end{eqnarray}
We can also orthogonal project $\delta f_i(\bm z)$ in $L^2(\rho_A\varphi_C)$ to separate the effects for active parts
\begin{eqnarray}
    \delta f_i^\parallel =P_A\delta f_i\equiv\int d\bm z_C\,\varphi_C\delta f_i,\label{eq:projection}
\end{eqnarray}
which only holds $\bm z_A$ dependent, and collapsed parts
\begin{eqnarray}
    \delta f_i^\perp=(1-P_A)\delta f_i.
\end{eqnarray}
Here, $P_A \delta f_i^\perp = 0$, and
\begin{eqnarray}
    \int d\bm z_A d\bm z_C \rho_A \varphi_C \delta f_i^\parallel \delta f_i^\perp = 0,
\end{eqnarray}
cancelling the couplings between $\delta \bm f^\parallel$ and $\delta \bm f^\perp$ in the Hessian~\eqref{eq:hessian_general}.
Moreover, the kernels with active encoder fields have no $\bm z_C$ dependence except $\delta f_i^\perp$, while
the kernels with collapsed encoder fields contain $z_c$ or $z_c^2-1$ whose Gaussian averages vanish.
Thus, the couplings of $\delta \bm f^\parallel$--$\delta\bm\mu_C$, $\delta \bm f^\parallel$--$\delta\bm\sigma_C$, $\delta \bm f^\perp$--$\delta\bm\mu_A$, and $\delta \bm f^\perp$--$\delta\bm\sigma_A$ also vanish.

We can rewrite $\delta\Phi$ as
\begin{eqnarray}
    \delta\Phi = (\delta\Phi_A, \delta\Phi_C),
\end{eqnarray}
where
\begin{eqnarray}
    \delta\Phi_A &:= (\delta\bm f^\parallel,\delta\bm\mu_A, \delta\bm\sigma_A),\\
    \delta\Phi_C &:= (\delta\bm f^\perp,\delta\bm\mu_C, \delta\bm\sigma_C),
\end{eqnarray}
the Hessian separates into the active and collapsed invariant fluctuation
sectors, represented by $\delta\Phi_A$ and $\delta\Phi_C$:
\begin{equation}
 \bm K=\bm K_A\oplus \bm K_C,\qquad Q=Q_A+Q_C.
 \label{eq:direct-sum}
\end{equation}
Substituting the partially collapsed background into the kernels gives:
\begin{align}
 Q_C&=\frac{1}{\sigma'^2}\int d\bm z_A\,d\bm z_C\,
 \rho_A(\bm z_A)\varphi_C(\bm z_C)\sum_i[\delta f_i^\perp(\bm z_A,\bm z_C)]^2\notag\\
 &\quad+\sum_{c\in C}\int d\bm x\,p_{\rm data}(\bm x)
       \big[(\delta\mu_c(\bm x))^2+2(\delta\sigma_c(\bm x))^2\big]\notag\\
 &\quad-\frac{2}{\sigma'^2}\sum_{c\in C}\sum_i\int d\bm x\,d\bm z_A\,d\bm z_C\,
 p_{\rm data}(\bm x)\bar q_A(\bm z_A|\bm x)\varphi_C(\bm z_C)\notag\\
 &\quad\quad\times r_i(\bm x,\bm z_A)\delta f_i^\perp(\bm z_A,\bm z_C)
       \big[z_c\delta\mu_c(\bm x)+(z_c^2-1)\delta\sigma_c(\bm x)\big].
 \label{eq:transverse-Q}
\end{align}

\subsection{Conditional residual operator and collapsed-sector stability}

We now analyze the collapsed-sector fluctuations $\delta\Phi_C$ through
$Q_C$ in Eq.~\eqref{eq:transverse-Q}, with the active background
$\bar f_i^A$, $\bar q_A$, and $\rho_A$ held fixed. The coupling to
$\delta\mu_c$ and $\delta\sigma_c$ involves only $z_c$ and $z_c^2-1$.
For each $c\in C$, introduce the normalized Hermite polynomials,
\begin{equation}
 h_1^c=z_c,\qquad
 h_2^c=\frac{z_c^2-1}{\sqrt2},
\end{equation}
which obey
\begin{align}
 \int d\bm z_C \varphi_C h_1^c h_1^d =\int d\bm z_C \varphi_C h_2^c h_2^d=\delta_{cd},
\end{align}
and,
\begin{eqnarray}
    \int d\bm z_C \varphi_C h_1^c h_2^d&=0.
\end{eqnarray}
Using the same projection $P_A$ as in Eq.~\eqref{eq:projection},
we can expand the collapsed-sector decoder fluctuation $\delta f_i^\perp$ in Hermite modes as
\begin{equation}
 \delta f_i^\perp
 =\sum_{c\in C}\left(h_1^c\delta f_{ic}^{(1)}+h_2^c\delta f_{ic}^{(2)}\right)
   +\delta f_i^{\rm rest},
 \label{eq:decoder-modes}
\end{equation}
where
\begin{align}
 \delta f_{ic}^{(1)}(\bm z_A)&=\int d\bm z_C \varphi_C
          h_1^c \delta f_i^\perp,\\
 \delta f_{ic}^{(2)}(\bm z_A)&=\int d\bm z_C\,\varphi_C
          h_2^c \delta f_i^\perp,
\end{align}
and the remainder $\delta f_i^{\rm rest}$ is orthogonal to $1,h_1^c,h_2^c$ at each $\bm z_A$.

The encoder perturbations and the decoder coefficients in
Eq.~\eqref{eq:decoder-modes} are functions of different variables.
For each $c\in C$, the scalar functions $\delta\mu_c(\bm x)$ and
$\delta\sigma_c(\bm x)$ belong to the real weighted Hilbert space
$\mathcal H_x=L^2(p_{\rm data})$, while the collections
$\{\delta f_{ic}^{(1)}(\bm z_A)\}_i$ and
$\{\delta f_{ic}^{(2)}(\bm z_A)\}_i$ belong to
$\mathcal H_A=L^2(\rho_A;\mathbb R^N)$.
For $h\in\mathcal H_x$ and $\bm g=(g_1,\ldots,g_N)\in\mathcal H_A$,
the corresponding squared norms are
\begin{align}
 \|h\|_x^2&=\langle h^2\rangle_{\bm x}=\int d\bm x\,p_{\rm data}(\bm x)h(\bm x)^2,\\
 \|\bm g\|_A^2&=\mathbb E_{\rho_A}\|\bm g\|^2=\sum_i\int d\bm z_A\,\rho_A(\bm z_A)g_i(\bm z_A)^2.
\end{align}
The inner products use the same integrals, with each square replaced
by the product of the two functions. For an individual decoder
coefficient or a remainder, we also use the scalar norms
\begin{align}
 \|a\|_{\rho_A}^2&=\int d\bm z_A\,\rho_A(\bm z_A)a(\bm z_A)^2,\\
 \|b\|_{\rho_A\varphi_C}^2&=\int d\bm z_A\,d\bm z_C\,
 \rho_A(\bm z_A)\varphi_C(\bm z_C)b(\bm z_A,\bm z_C)^2.
\end{align}

The mixed terms in $Q_C$ couple these two spaces through the
residual field $r_i$. To express this coupling, define the
conditional residual operator $\mathcal R_A:\mathcal H_x\to\mathcal H_A$ by
\begin{equation}
 (\mathcal R_A h)_i(\bm z_A)
 =\frac{\int d\bm x\,p_{\rm data}(\bm x)\bar q_A(\bm z_A|\bm x)
                                  r_i(\bm x,\bm z_A)h(\bm x)}
             {\rho_A(\bm z_A)}.
 \label{eq:residual-operator}
\end{equation}
Equivalently, $(\mathcal R_Ah)_i(\bm z_A)=
\mathbb E[r_i(\bm x,\bm z_A)h(\bm x)\mid\bm z_A]$,
with expectation under $p_{\rm data}(\bm x)\bar q_A(\bm z_A|\bm x)$,
as in the main text. Thus $\mathcal R_A$ takes a scalar encoder perturbation, such as
$\delta\mu_c$ or $\delta\sigma_c$, to an $N$-component function of
$\bm z_A$ as the decoder coefficients. The definition
is understood on the support of $\rho_A$ and for functions for which
the displayed integrals are finite. The factor $\rho_A^{-1}$ converts
the coupling integral into the weighted inner product in $\mathcal H_A$.

The adjoint $\mathcal R_A^\dagger:\mathcal H_A\to\mathcal H_x$ is
determined by transferring $\mathcal R_A$ from one side of this inner
product to the other. Substituting Eq.~\eqref{eq:residual-operator}
and interchanging the integrations gives
\begin{align}
 &\sum_i\int d\bm z_A\,\rho_A(\bm z_A)
                 g_i(\bm z_A)(\mathcal R_Ah)_i(\bm z_A)\notag\\
 &\quad=\int d\bm x\,p_{\rm data}(\bm x)h(\bm x)
       \left[\sum_i\int d\bm z_A\,\bar q_A(\bm z_A|\bm x)
                  r_i(\bm x,\bm z_A)g_i(\bm z_A)\right]\notag\\
 &\quad=\int d\bm x\,p_{\rm data}(\bm x)h(\bm x)
                             (\mathcal R_A^\dagger\bm g)(\bm x).
 \label{eq:residual-adjoint-identity}
\end{align}
Comparing the last two lines identifies the adjoint as
\begin{equation}
 (\mathcal R_A^\dagger\bm g)(\bm x)
 =\sum_i\int d\bm z_A\,\bar q_A(\bm z_A|\bm x)
                  r_i(\bm x,\bm z_A)g_i(\bm z_A),
 \label{eq:residual-adjoint}
\end{equation}
which maps the decoder coefficients back to a scalar function of $\bm x$.

Equation~\eqref{eq:transverse-Q} now becomes
\begin{align}
 Q_C={}&\frac{1}{\sigma'^2}\sum_i\int d\bm z_A\,d\bm z_C\,
       \rho_A(\bm z_A)\varphi_C(\bm z_C)[\delta f_i^{\rm rest}]^2\notag\\
 &+\sum_{c\in C}\int d\bm x\,p_{\rm data}(\bm x)
                   \big[(\delta\mu_c)^2+2(\delta\sigma_c)^2\big]\notag\\
 &+\frac{1}{\sigma'^2}\sum_{c\in C}\sum_i\int d\bm z_A\,\rho_A(\bm z_A)
       \Big[(\delta f_{ic}^{(1)})^2+(\delta f_{ic}^{(2)})^2\notag\\
 &\hspace{28mm}-2\delta f_{ic}^{(1)}(\mathcal R_A\delta\mu_c)_i
                   -2\sqrt2\delta f_{ic}^{(2)}(\mathcal R_A\delta\sigma_c)_i\Big].
 \label{eq:operator-Q}
\end{align}
Thus each collapsed coordinate contributes independent posterior-mean
and posterior-width mass blocks. We parameterize the latter by the
standard-deviation fluctuation $\delta\sigma_c$, with
$\delta(\sigma_c^2)=2\delta\sigma_c$ at the collapsed background.
These blocks act respectively on
$(\{\delta f_{ic}^{(1)}\}_i,\delta\mu_c)$ and
$(\{\delta f_{ic}^{(2)}\}_i,\delta\sigma_c)$:
\begin{align}
 \bm K_{C,\mu}&=\begin{pmatrix}\sigma'^{-2}I&-\sigma'^{-2}\mathcal R_A\\
 -\sigma'^{-2}\mathcal R_A^\dagger&I\end{pmatrix},\\
 \bm K_{C,\sigma}&=\begin{pmatrix}\sigma'^{-2}I&-\sqrt2\sigma'^{-2}\mathcal R_A\\
 -\sqrt2\sigma'^{-2}\mathcal R_A^\dagger&2I\end{pmatrix}.
 \label{eq:operator-blocks}
\end{align}
Here the matrices represent the mass operators induced by $Q_C$
in the weighted function spaces above, after integration over the
collapsed Gaussian variables. They are not pointwise matrices at a
fixed $(\bm x,\bm z)$. The remainder $\delta f_i^{\rm rest}$ contributes
the positive mass block $\sigma'^{-2}I$ to $\bm K_C$.
Thus these remaining modes are strictly massive, and the collapsed sector decomposes as
\begin{equation}
 \bm K_C\simeq
 \bigoplus_{c\in C}\big(\bm K_{C,\mu}\oplus\bm K_{C,\sigma}\big)
 \oplus\sigma'^{-2}I_{\rm rest}.
\end{equation}

Define the positive semi-definite Gram operator
\begin{equation}
 \mathcal C_A=\mathcal R_A^\dagger\mathcal R_A,
 \label{eq:C}
\end{equation}
whose spectrum is the residual spectrum of the partially collapsed
representation. Completing squares, or taking the Schur complement of the decoder
blocks, gives the exact identity
\begin{align}
 Q_C={}&\frac{1}{\sigma'^2}\sum_i\int d\bm z_A\,d\bm z_C\,
       \rho_A(\bm z_A)\varphi_C(\bm z_C)[\delta f_i^{\rm rest}]^2\notag\\
 &+\frac{1}{\sigma'^2}\sum_{c\in C}\sum_i\int d\bm z_A\,\rho_A(\bm z_A)
       \big[\delta f_{ic}^{(1)}-(\mathcal R_A\delta\mu_c)_i\big]^2\notag\\
 &+\frac{1}{\sigma'^2}\sum_{c\in C}\sum_i\int d\bm z_A\,\rho_A(\bm z_A)
       \big[\delta f_{ic}^{(2)}-\sqrt2(\mathcal R_A\delta\sigma_c)_i\big]^2\notag\\
 &+\sum_{c\in C}\int d\bm x\,p_{\rm data}(\bm x)
       \Big[\delta\mu_c(\bm x)[(I-\mathcal C_A/\sigma'^2)\delta\mu_c](\bm x)\notag\\
 &\hspace{30mm}+2\delta\sigma_c(\bm x)[(I-\mathcal C_A/\sigma'^2)\delta\sigma_c](\bm x)\Big].
 \label{eq:squares}
\end{align}
Assume $\mathcal R_A$ is bounded. The leading residual eigenvalue
(more generally, the supremum of the residual spectrum) is
\begin{equation}
 \Lambda_*^{(A)}=\|\mathcal R_A\|^2
 =\sup_{h\ne0}\sum_i \frac{\int d\bm z_A\,\rho_A(\bm z_A)[(\mathcal R_Ah)_i(\bm z_A)]^2}{\int d\bm x\,p_{\rm data}(\bm x)h(\bm x)^2}
 =\sup\spec\mathcal C_A.
\end{equation}
The necessary and sufficient condition for nonnegative curvature in
the collapsed sector is therefore
\begin{equation}
 \boxed{\bm K_C\succeq0\quad\Longleftrightarrow\quad
                  \sigma'^2\geq\Lambda_*^{(A)}.}
 \label{eq:criterion}
\end{equation}

For the complete stationary solution, one must also require
$\bm K_A\succeq0$. Since $Q=Q_A+Q_C$, the collapsed perturbations can be
analyzed independently of $\delta\Phi_A$ at quadratic order. Their
stability still depends on the active background through $\mathcal R_A$.

\subsection{Mass spectrum and soft modes}


For a positive residual eigenvalue, choose
$\mathcal C_A\psi_\alpha=\Lambda_\alpha^{(A)}\psi_\alpha$ with
$\langle\psi_\alpha^2\rangle_{\bm x}=1$, and define its decoder partner by
\begin{equation}
 \bm\chi_\alpha=\frac{\mathcal R_A\psi_\alpha}
                         {\sqrt{\Lambda_\alpha^{(A)}}}.
\end{equation}
Then $\mathbb E_{\rho_A}\|\bm\chi_\alpha\|^2=1$ and
$\mathcal R_A^\dagger\bm\chi_\alpha
 =\sqrt{\Lambda_\alpha^{(A)}}\psi_\alpha$.
The two-dimensional subspace
\begin{equation}
 \mathcal V_\alpha=\operatorname{span}\big\{
 (\bm\chi_\alpha,0),(0,\psi_\alpha)\big\}
\end{equation}
is invariant under each mass operator. Its first component is the
first Hermite decoder coefficient for the mean channel and the second
Hermite coefficient for the width channel. In this orthonormal basis,
the restrictions of the operators have the matrix representations
\begin{align}
 \left.\bm K_{C,\mu}\right|_{\mathcal V_\alpha}
 &\widehat{=}\bm M_{\mu,\alpha}:=
 \begin{pmatrix}
  1/\sigma'^2&-\sqrt{\Lambda_\alpha^{(A)}}/\sigma'^2\\
  -\sqrt{\Lambda_\alpha^{(A)}}/\sigma'^2&1
 \end{pmatrix},\notag\\
 \left.\bm K_{C,\sigma}\right|_{\mathcal V_\alpha}
 &\widehat{=}\bm M_{\sigma,\alpha}:=
 \begin{pmatrix}
  1/\sigma'^2&-\sqrt{2\Lambda_\alpha^{(A)}}/\sigma'^2\\
  -\sqrt{2\Lambda_\alpha^{(A)}}/\sigma'^2&2
 \end{pmatrix}.
 \label{eq:restricted-mass-blocks}
\end{align}
Here $\widehat{=}$ denotes a matrix representation on $\mathcal V_\alpha$;
$\bm K_{C,\mu}$ and $\bm K_{C,\sigma}$ retain their operator meaning.
The matrix $\bm M_{\mu,\alpha}$ is the mean-channel block used in the main text;
$\bm M_{\sigma,\alpha}$ denotes the corresponding width-channel block.
For a single collapsed coordinate, the mode amplitudes $u,v$ in the
main text give $\delta\bm f^\perp=u z_c\bm\chi_\alpha$ and
$\delta\mu_c=v\psi_\alpha$, with quadratic free-energy contribution
$\tfrac12(u,v)\bm M_{\mu,\alpha}(u,v)^{\mathsf T}$.
Diagonalization gives
\begin{align}
 m^2_{\mu,\alpha,\pm}
 &=\frac{1+\sigma'^2\pm\sqrt{(1-\sigma'^2)^2+4\Lambda_\alpha^{(A)}}}{2\sigma'^2},\\
 m^2_{\sigma,\alpha,\pm}
 &=\frac{1+2\sigma'^2\pm\sqrt{(1-2\sigma'^2)^2+8\Lambda_\alpha^{(A)}}}{2\sigma'^2}.
 \label{eq:masses}
\end{align}
Both traces are positive, while the determinants are respectively
$(\sigma'^2-\Lambda_\alpha^{(A)})/\sigma'^4$ and
$2(\sigma'^2-\Lambda_\alpha^{(A)})/\sigma'^4$.
Thus both lower branches vanish at
$\sigma'^2=\Lambda_\alpha^{(A)}$ and become negative below it.
At equality, their zero-mode amplitudes obey
\begin{align}
 \delta\bm f_c^{(1)}&=\mathcal R_A\delta\mu_c,\\
 \delta\bm f_c^{(2)}&=\sqrt2\mathcal R_A\delta\sigma_c,
\end{align}
where $\delta\bm f_c^{(k)}:=\{\delta f_{ic}^{(k)}\}_{i=1}^N$
for $k=1,2$, and the encoder perturbation is proportional to $\psi_\alpha$.
These relations also follow from the completed squares in
Eq.~\eqref{eq:squares}. Each mass branch is repeated $|C|$ times,
in addition to the multiplicity of the residual eigenvalue.

Encoder modes in $\ker\mathcal R_A$ have squared masses $1$ and $2$
in the mean and width channels, respectively. Decoder modes in
$\ker\mathcal R_A^\dagger$, together with the remaining Hermite modes,
have squared mass $\sigma'^{-2}$. Hence the stability boundary is
controlled by the upper edge of the residual spectrum. If that edge
is attained by an eigenmode, marginality produces the zero modes
above; otherwise it marks closure of the spectral gap without a
normalizable eigenmode at zero.

\subsection{Interpretation and the fully collapsed limit}

For the active background, Eq.~\eqref{eq:decoder-stationary} gives a conditional expectation of $x_i$ under the distribution $p_{\rm data}(\bm x) \bar q_A(\bm z_A| \bm x)$:
\begin{equation}
 \bar f_i^A(\bm z_A)=
 \frac{\int d\bm x\,p_{\rm data}(\bm x)\bar q_A(\bm z_A|\bm x)x_i}
      {\rho_A(\bm z_A)}.
\end{equation}
Consequently, the residual operator can be written as
\begin{align}
 (\mathcal R_Ah)_i(\bm z_A)
 ={}&\frac{\int d\bm x\,p_{\rm data}(\bm x)\bar q_A(\bm z_A|\bm x)x_i h(\bm x)}
          {\rho_A(\bm z_A)}\notag\\
 &-\bar f_i^A(\bm z_A)
 \frac{\int d\bm x\,p_{\rm data}(\bm x)\bar q_A(\bm z_A|\bm x)h(\bm x)}
      {\rho_A(\bm z_A)},
\end{align}
which is the conditional covariance of $x_i$ and $h(\bm x)$ at fixed
$\bm z_A$ under $p_{\rm data}(\bm x)\bar q_A(\bm z_A|\bm x)$. In particular, $\mathcal R_A1=0$, and
\begin{equation}
 \Lambda_*^{(A)}=\sup_{\langle h^2\rangle_{\bm x}=1}\int d\bm z_A\,\rho_A(\bm z_A)\sum_i
 \left[
 \frac{\int d\bm x\,p_{\rm data}(\bm x)\bar q_A(\bm z_A|\bm x)
                 [x_i-\bar f_i^A(\bm z_A)]h(\bm x)}
      {\rho_A(\bm z_A)}\right]^2.
 \label{eq:rayleigh}
\end{equation}
The leading eigenvalue of the data covariance matrix admits the same
function-space variational formulation.
For finite data covariance $\bm\Sigma_X$,
\begin{equation}
 \lambda_{\rm max}(\bm\Sigma_X)
 =\sup_{\langle h^2\rangle_{\bm x}=1}\sum_i
 \left[\int d\bm x\,p_{\rm data}(\bm x)
                  (x_i-\langle x_i\rangle_{\bm x})h(\bm x)\right]^2,
 \label{eq:pca-leading}
\end{equation}
gives the leading covariance eigenvalue~\cite{blanchard2007statistical}.
Analogously, the \textit{leading residual eigenvalue} $\Lambda_*^{(A)}$
measures the strongest residual correlation accessible to an additional
latent degree of freedom given the active representation.

For a fully collapsed stationary state, $A=\varnothing$, we have
\begin{equation}
 (\mathcal R_\varnothing h)_i=\int d\bm x\,p_{\rm data}(\bm x)(x_i-\langle x_i\rangle_{\bm x})h(\bm x),\qquad \Lambda_*^{(\varnothing)} = \lambda_{\rm max}(\bm\Sigma_X).
\end{equation}
This recovers the full-collapse threshold in Ref.~\cite{li2026posterior}.

\subsection{Linear perturbations and the linear VAE}

We evaluate the residual spectrum for linear fluctuations directly at
a partially collapsed stationary state of the linear Gaussian VAE.
Let the data be Gaussian with mean $\langle\bm x\rangle_{\bm x}$ and
covariance $\bm\Sigma_X$. Denote its positive eigenvalues and
orthonormal eigenvectors by $\lambda_\alpha=\lambda_\alpha(\bm\Sigma_X)$
and $\bm e_\alpha$, with $\lambda_1\geq\lambda_2\geq\cdots>0$, and write
\begin{equation}
 y_\alpha(\bm x):=\sum_i e_{\alpha i}
       (x_i-\langle x_i\rangle_{\bm x}),\qquad
 \langle y_\alpha y_\beta\rangle_{\bm x}
       =\lambda_\alpha\delta_{\alpha\beta}.
\end{equation}
All data directions below are restricted to the support of
$\bm\Sigma_X$. We label each active latent coordinate by the principal
direction it represents, so that $A$ also labels the represented
principal directions, as in the main text. The notation
$\alpha\notin A$ refers to unrepresented data directions; $c\in C$
continues to label collapsed latent coordinates.

With learned encoder variances, the stationary solution is~\cite{dang2024beyond}
\begin{align}
 \bar f_i^A(\bm z_A)&=\langle x_i\rangle_{\bm x}
       +\sum_{a\in A}\sqrt{\lambda_a-\sigma'^2}\,e_{ai}z_a,\\
 \bar\mu_a(\bm x)&=\frac{\sqrt{\lambda_a-\sigma'^2}}{\lambda_a} y_a(\bm x),\\
 \bar\sigma_a^2&=\frac{\sigma'^2}{\lambda_a},
 \label{eq:linear-stationary-branch}
\end{align}
where $\lambda_a>\sigma'^2$, while $\bar\mu_c=0$ and
$\bar\sigma_c=1$ for $c\in C$. Under the joint distribution
$p_{\rm data}(\bm x)\bar q_A(\bm z_A|\bm x)$, the active latent
marginal $\rho_A$ is standard Gaussian. Gaussian conditioning gives
\begin{align}
 \mathbb E[y_\alpha\mid\bm z_A]
 &=\begin{cases}
   \sqrt{\lambda_\alpha-\sigma'^2}\,z_\alpha,&\alpha\in A,\\
   0,&\alpha\notin A,
   \end{cases}\\
 \operatorname{Cov}(y_\alpha,y_\beta\mid\bm z_A)
 &=\delta_{\alpha\beta}
   \begin{cases}
   \sigma'^2,&\alpha\in A,\\
   \lambda_\alpha,&\alpha\notin A.
   \end{cases}
 \label{eq:linear-conditional-covariance}
\end{align}
In particular, $\bar f_i^A=\mathbb E[x_i\mid\bm z_A]$, so the residual
field is the same $r_i=x_i-\bar f_i^A$ used above.

To evaluate the conditional residual operator, choose the normalized
linear encoder modes and constant decoder modes
\begin{equation}
 \psi_\alpha(\bm x)=\frac{y_\alpha(\bm x)}{\sqrt{\lambda_\alpha}},
 \qquad \bm\chi_\alpha(\bm z_A)=\bm e_\alpha.
 \label{eq:linear-modes}
\end{equation}
They are orthonormal in $\mathcal H_x$ and $\mathcal H_A$, respectively.
Substituting Eq.~\eqref{eq:linear-conditional-covariance} into the
conditional residual operator gives
\begin{align}
 (\mathcal R_A\psi_\alpha)_i
 &=\frac{\mathbb E[r_i y_\alpha\mid\bm z_A]}{\sqrt{\lambda_\alpha}}
   =\sqrt{\Lambda_\alpha^{(A)}}\chi_{\alpha i},\\
 \sqrt{\Lambda_\alpha^{(A)}}&=
 \begin{cases}
   \sigma'^2/\sqrt{\lambda_\alpha},&\alpha\in A,\\
   \sqrt{\lambda_\alpha},&\alpha\notin A.
 \end{cases}
 \label{eq:linear-residual-operator}
\end{align}
The adjoint in Eq.~\eqref{eq:residual-adjoint} similarly gives
\begin{align}
 (\mathcal R_A^\dagger\bm\chi_\alpha)(\bm x)
 &=\sum_i e_{\alpha i}\int d\bm z_A\,
       \bar q_A(\bm z_A|\bm x)r_i(\bm x,\bm z_A)\notag\\
 &=\begin{cases}
   (\sigma'^2/\lambda_\alpha)y_\alpha(\bm x),&\alpha\in A,\\
   y_\alpha(\bm x),&\alpha\notin A
   \end{cases}\\
   &=\sqrt{\Lambda_\alpha^{(A)}}\psi_\alpha(\bm x).
 \label{eq:linear-residual-adjoint}
\end{align}
Thus these modes form singular pairs of $\mathcal R_A$ itself, and
$\mathcal C_A\psi_\alpha=\Lambda_\alpha^{(A)}\psi_\alpha$ with
\begin{equation}
 \Lambda_\alpha^{(A)}=
 \begin{cases}
  \sigma'^4/\lambda_\alpha,&\alpha\in A,\\
  \lambda_\alpha,&\alpha\notin A.
 \end{cases}
 \label{eq:linear-pca-spectrum}
\end{equation}
Here $\alpha$ indexes the linear modes in Eq.~\eqref{eq:linear-modes};
this is the residual spectrum relevant to linear fluctuations, not a
claim that these modes exhaust the unrestricted functional spectrum.

In the collapsed sector, linear decoder and encoder-mean fluctuations
can now be expanded using the same singular pairs:
\begin{align}
 \delta f_i^\perp(\bm z)
 &=\sum_{c\in C}\sum_\alpha
       z_c\,\delta f_{\alpha c}^{(1)}\chi_{\alpha i},\\
 \delta\mu_c(\bm x)&=\sum_\alpha\delta\mu_{\alpha c}\psi_\alpha(\bm x).
 \label{eq:linear-perturbations}
\end{align}
The coefficients $\delta f_{\alpha c}^{(1)}$ and
$\delta\mu_{\alpha c}$ are constants. A constant encoder-mean
fluctuation decouples because $\mathcal R_A1=0$ and has squared mass
one. Since a linear decoder has no second Hermite mode,
$\delta f_{ic}^{(2)}=\delta f_i^{\rm rest}=0$.
The posterior-width fluctuations therefore decouple and contribute
$2\sum_{c\in C}\|\delta\sigma_c\|_x^2$, even if
$\delta\sigma_c$ is allowed to be an arbitrary function of $\bm x$.
Equation~\eqref{eq:operator-Q} reduces to
\begin{align}
 Q_C^{\rm lin}
 ={}&\sum_{c\in C}\sum_\alpha
 \left[
 \frac{(\delta f_{\alpha c}^{(1)}-\sqrt{\Lambda_\alpha^{(A)}}\delta\mu_{\alpha c})^2}
      {\sigma'^2}
 +\left(1-\frac{\Lambda_\alpha^{(A)}}{\sigma'^2}\right)
      (\delta\mu_{\alpha c})^2
 \right]\notag\\
 &+2\sum_{c\in C}\|\delta\sigma_c\|_x^2.
 \label{eq:linear-squares}
\end{align}
The mean-sector matrices are $\bm M_{\mu,\alpha}$, representing the restrictions of $\bm K_{C,\mu}$ in Eq.~\eqref{eq:restricted-mass-blocks},
with squared masses given by Eq.~\eqref{eq:masses} and residual
eigenvalues given by Eq.~\eqref{eq:linear-pca-spectrum}.
For a nonempty collapsed sector, nonnegative curvature under these
linear fluctuations is equivalent to
$\sigma'^2\geq\max_\alpha\Lambda_\alpha^{(A)}$.
Since $\lambda_a>\sigma'^2$ on the active stationary branch,
represented directions satisfy $\sigma'^4/\lambda_a<\sigma'^2$.
The stability criterion therefore reduces to
\begin{equation}
 \sigma'^2\geq\max_{\alpha\notin A}\lambda_\alpha,
 \label{eq:linear-criterion}
\end{equation}
where the maximum is taken to be zero if no unrepresented direction
remains. This recovers the principal component thresholds of the
linear Gaussian VAE discussed in the main text and in Appendix~D.2
of Ref.~\cite{dang2024beyond}. For $A=\varnothing$, it reduces to the
full-collapse threshold $\lambda_{\rm max}(\bm\Sigma_X)$.

\subsection{Simulation details and additional results}
\label{sec:simulation-details}

(Codes are available at \href{https://github.com/lizhen0114/vae-residual-spectral-instability}{here})

\subsubsection{Data construction}

The experiment uses an equally weighted, covariance-matched mixture of two
Gaussian distributions in $N=8$ dimensions. Let $\bm G\sim\mathcal N(0,I_8)$
and let $\eta\in\{-1,+1\}$ take either value with probability $1/2$,
independently of $\bm G$. Define
\begin{align}
 \bm v&=\frac{1}{\sqrt8}(1,\ldots,1)^{\mathsf T},\qquad a=0.5,\\
 \bm Y&=(I_8-a^2\bm v\bm v^{\mathsf T})^{1/2}\bm G+a\eta\bm v,\\
 \bm X&=O D^{1/2}\bm Y,\qquad
 D=\operatorname{diag}(\lambda_1,\ldots,\lambda_8),\\
 \lambda_i&=\exp[-0.1(i-1)],\qquad i=1,\ldots,8.
 \label{eq:mixture-construction}
\end{align}
Here $O$ is a fixed orthogonal matrix obtained by QR decomposition of an
$8\times8$ matrix of independent standard Gaussian entries.
Conditioned on $\eta$, the two components have means
$\pm a O D^{1/2}\bm v$ and common covariance
$O D^{1/2}(I_8-a^2\bm v\bm v^{\mathsf T})D^{1/2}O^{\mathsf T}$.
The mixture therefore satisfies
\begin{equation}
 \mathbb E[\bm X]=0,\qquad \operatorname{Cov}(\bm X)=O D O^{\mathsf T}.
\end{equation}
This construction changes higher-order statistics while preserving the
population covariance spectrum. In particular,
$U=\bm v^{\mathsf T}\bm Y=\sqrt{1-a^2}G_\parallel+a\eta$ has unit variance
and excess kurtosis $-2a^4=-0.125$. At $a=0$, the distribution reduces to
the Gaussian covariance-matched reference.

We generate $256000$ samples using data seed $1234$ and subtract their
sample mean. The covariance is left at its sampled value, so its empirical
eigenvalues differ slightly from the prescribed population values
$\lambda_i$. The same centered dataset is used throughout training and
spectral estimation.
\subsubsection{Network and continuation protocol}

The latent dimension is $n=4$. The encoder has hidden widths $256,128$
with ReLU activations and separate linear output heads for $\bm\mu$ and
$\log\bm\sigma^2$. The log variances are clamped to $[-12,8]$.
The decoder has hidden widths $128,256$ with ReLU activations and an
$8$-dimensional linear output. The prior is standard Gaussian and the
isotropic decoder variance $\sigma'^2$ is fixed at each continuation point.
Training minimizes the empirical negative ELBO in Eq.~\eqref{eq:free-energy},
using one reparameterized latent sample per input and the analytical
Gaussian KL term.

The variance increases from $0.65$ to $1.05$ in increments of $0.001$,
giving $401$ checkpoints. With training seed $0$, the first point is
trained for $120$ epochs; each subsequent point is initialized from the
previous network weights and trained for $30$ epochs. 

At each checkpoint, we evaluate the per-coordinate mean KL over the full
dataset,
\begin{equation}
 \overline{\mathrm{KL}}_j=
 \frac12\left\langle\mu_j^2+\sigma_j^2-1-\log\sigma_j^2
 \right\rangle_{\bm x}.
\end{equation}
A coordinate belongs to $A$ when $\overline{\mathrm{KL}}_j>10^{-3}$;
the remaining coordinates form $C$.
This finite-threshold classification provides an operational approximation
to the exact collapsed background used in the theory. The numerical
results here use one training seed; repeated spectral measurements below
quantify sampling variability at fixed network weights.

\subsubsection{Measurement of the residual spectral upper edge}

Spectral estimation is performed after training, with every checkpoint
frozen. We measure checkpoints with $C\ne\varnothing$, including complete
collapse. In the following, $\bar q_A$ denotes the
active encoder extracted from the frozen network.

\textit{Sampling and decoder projection.} For one estimator repeat, draw
$B$ data points $\{\bm x_b\}_{b=1}^B$ uniformly without replacement from
the training set. Draw $L$ active latent samples by first selecting a data
index uniformly, with replacement, and then sampling its active posterior.
Thus the sampling density is
\begin{equation}
 \widehat\rho_A(\bm z_A)=\frac1B\sum_{b=1}^B
                          \bar q_A(\bm z_A\mid\bm x_b).
\end{equation}
For each sampled $\bm z_{A,\ell}$, approximate the projected decoder by
\begin{equation}
 \widehat{\bm f}^{A}(\bm z_{A,\ell})=
 \frac{1}{K_{\rm MC}}\sum_{k=1}^{K_{\rm MC}}
 \bm f_\theta(\bm z_{A,\ell},\bm\xi_{C,k}),
 \quad \bm\xi_{C,k}\sim\mathcal N(0,I_{|C|}),\quad K_{\rm MC}=128.
 \label{eq:numerical-decoder-projection}
\end{equation}
Within a repeat, the same collapsed-coordinate draws are used at every
active latent sample. Decoder outputs are averaged before constructing
residuals. For $A=\varnothing$, a single projected decoder value is shared
by all quadrature rows. Projection approximates $P_A\bm f_\theta$; its
identification with the conditional mean relies on the stationary-background
assumption.

\begin{figure}[b]
    \centering
    \includegraphics[width=0.9\linewidth]{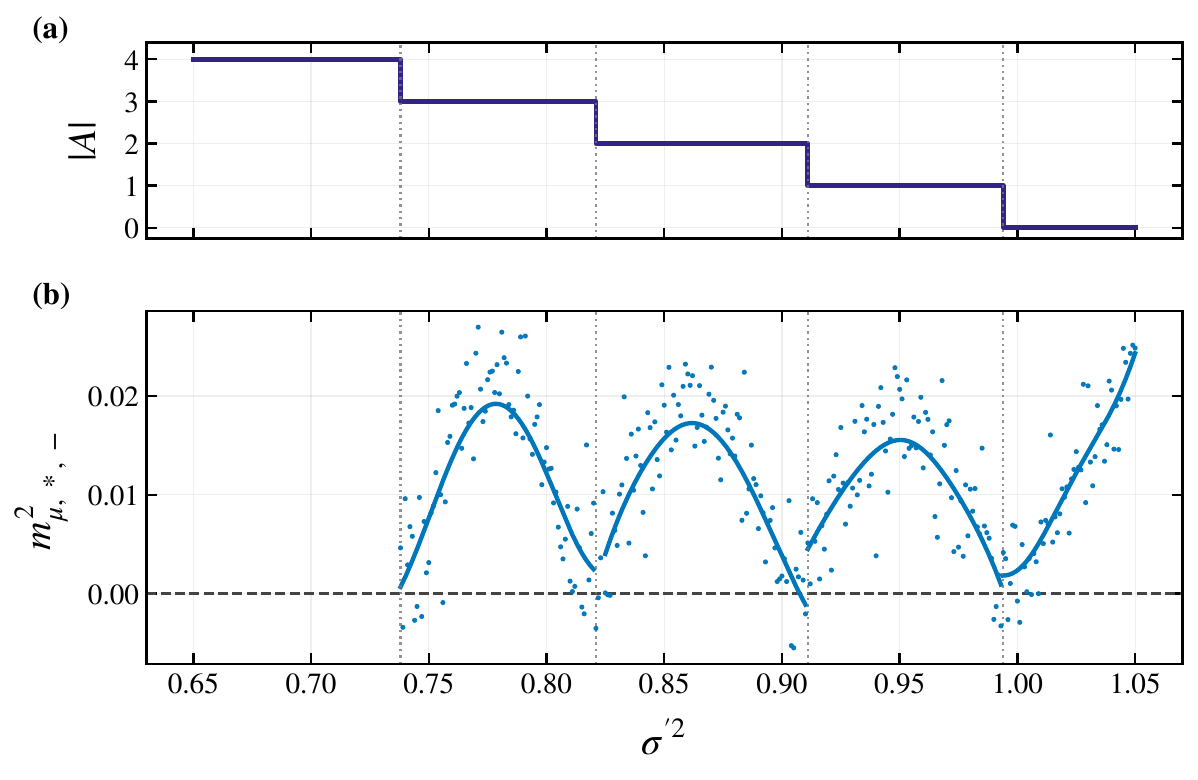}
    \caption{Additional results for the covariance-matched Gaussian mixture
    with $a=0.5$ and training seed $0$.
    (a) Number of active latent coordinates $|A|$ versus decoder variance
    $\sigma'^2$.
    (b) Squared mass $m^2_{\mu,*,-}$ obtained from the estimated leading
    residual eigenvalue using Eq.~\eqref{eq:masses}.
    Dots show the checkpoint estimates; solid curves are segmented
    Gaussian local-linear smooth guides. Vertical dotted lines mark
    the displayed changes in $|A|$, and the horizontal dashed line
    marks zero mass. Panel (b) is evaluated where the collapsed sector
    is nonempty.}
    \label{fig:additional-masses}
\end{figure}

\textit{Discrete residual operator.} Normalize posterior weights over the
sampled data and define
\begin{align}
 w_{\ell b}&=\frac{\bar q_A(\bm z_{A,\ell}\mid\bm x_b)}
                   {\sum_{b'=1}^B\bar q_A(\bm z_{A,\ell}\mid\bm x_{b'})},\\
 r_{\ell ib}&=x_{bi}-\widehat f_i^A(\bm z_{A,\ell}),\qquad
 (T_\ell)_{ib}=w_{\ell b}r_{\ell ib}.
\end{align}
The weights are evaluated by a softmax of log posterior densities.
For a vector of sampled function values $\bm h=(h(\bm x_1),\ldots,h(\bm x_B))^{\mathsf T}$,
$T_\ell\bm h$ approximates $(\mathcal R_Ah)(\bm z_{A,\ell})$.
Using data norm $B^{-1}\bm h^{\mathsf T}\bm h$ and quadrature weight $L^{-1}$
gives the empirical Gram operator
\begin{equation}
 \widehat{\mathcal C}_{A;B,L}
 =\frac{B}{L}\sum_{\ell=1}^L T_\ell^{\mathsf T}T_\ell.
 \label{eq:empirical-residual-gram}
\end{equation}

Power iteration is restricted to centered vectors with the projector
$P_0=I_B-\bm1\bm1^{\mathsf T}/B$. Starting from a centered, normalized
Gaussian random vector, each step applies the Gram operator, subtracts
the mean, and normalizes the resulting vector. The estimate is the Rayleigh quotient
$\widehat\Lambda(B,L)=\bm h^{\mathsf T}\widehat{\mathcal C}_{A;B,L}\bm h$
for $\bm h^{\mathsf T}\bm h=1$. Iteration stops after at most $100$ steps,
or when successive quotients differ by at most
$10^{-6}\max(1,|\widehat\Lambda_{\rm previous}|)$.
Centering implements the theoretical identity $\mathcal R_A1=0$.
The code also records the empirical conditional residual mean
$\sum_b w_{\ell b}\bm r_{\ell b}$ as a diagnostic of stationarity and
sampling error.

\textit{Sampling-bias extrapolation and uncertainty.} At each checkpoint,
we use the nine combinations
\begin{equation}
 B\in\{1024,2048,4096\},\qquad L\in\{256,512,1024\},
\end{equation}
with $16$ independently seeded estimator repeats per combination.
We fit the cell means to the empirical extrapolation model
\begin{equation}
 \widehat\Lambda(B,L)=\Lambda_\infty+\frac{b}{B}+\frac{c}{L},
 \label{eq:spectral-extrapolation}
\end{equation}
using inverse squared standard errors of the cell means as weights.
The fitted intercept defines the reported $\hat\Lambda_*^{(A)}$.

\subsubsection{Additional results}

Figure~\ref{fig:additional-masses} shows the mean-channel squared mass
associated with the estimated residual spectral upper edge along the
continuation trajectory. We evaluate $m^2_{\mu,*,-}$ from
Eq.~\eqref{eq:masses} by substituting $\hat\Lambda_*^{(A)}$ for
$\Lambda_\alpha^{(A)}$. This quantity expresses the measured residual
spectrum in the field-space mass normalization used in the main text.
The estimates approach zero near successive decreases of $|A|$ and
increase within the intervening branches. The scatter displays the
sampling variability of the spectral estimates near marginality.

\label{sec:direct-curvature}

\begin{figure}
    \centering
    \includegraphics[width=\linewidth]{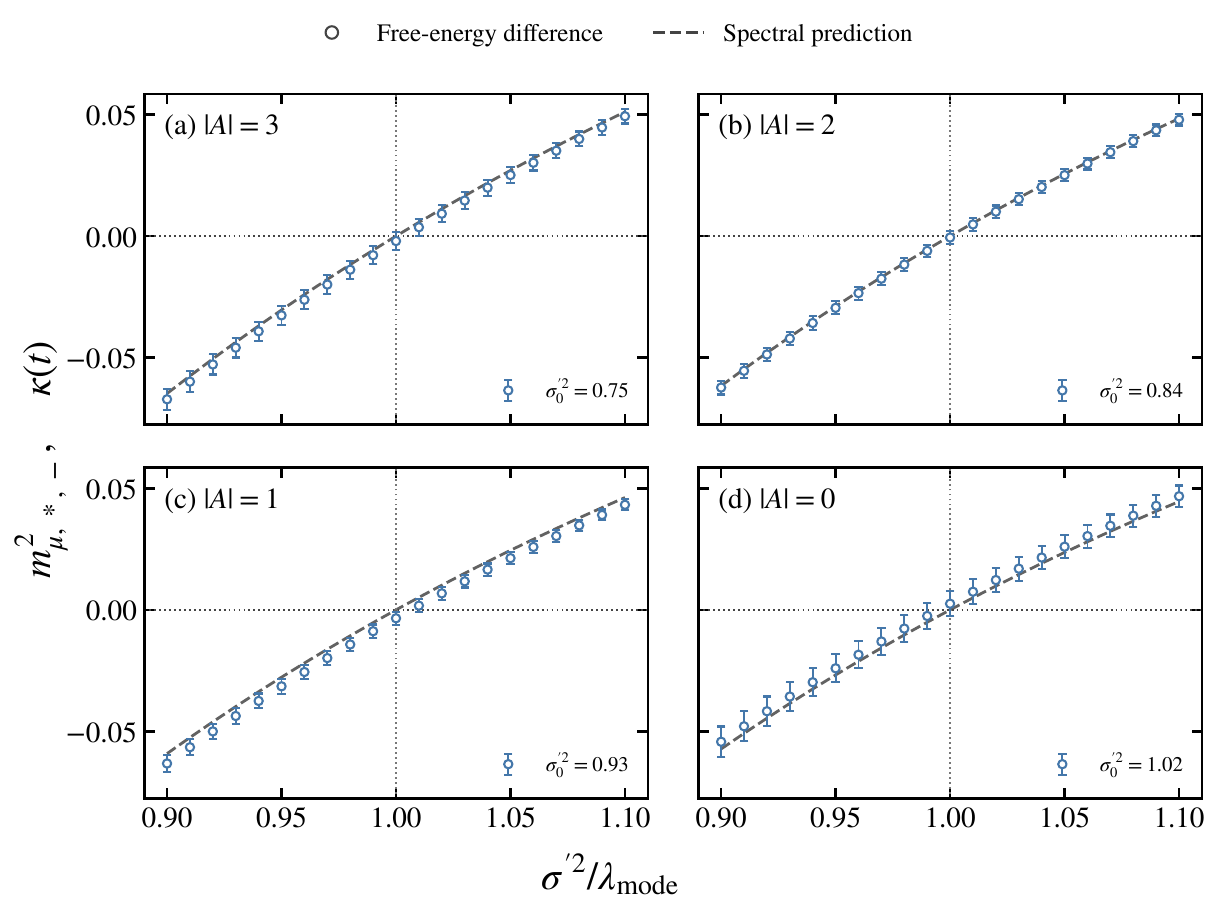}
    \caption{Direct free-energy curvature checks for fixed
    projected backgrounds with (a) $|A|=3$, (b) $|A|=2$, (c) $|A|=1$,
    and (d) $|A|=0$. Colors and marker shapes distinguish the checkpoint
    variances $\sigma_0'^2$ given in each legend. Points show
    $\kappa(0.01)$ from Eq.~\eqref{eq:numerical-direct-curvature}; error
    bars denote one standard error over four measurement repeats.
    Dashed curves show the predicted lower mass evaluated with
    $\lambda_{\rm mode}$. Dotted lines mark zero curvature and the
    predicted marginal ratio $\sigma'^2/\lambda_{\rm mode}=1$.
    Measured points are displayed without smoothing or a zero constraint.
    Asterisks identify backgrounds containing repeats that reached the
    iteration limit before satisfying the eigenvector-residual tolerance;
    these repeats are retained. The scans vary the decoder variance
    at fixed background fields.}
    \label{fig:direct-curvature}
\end{figure}

We additionally test the mean-channel fluctuation prediction by evaluating
free-energy changes along explicitly constructed coupled modes.
For the mixture with $a=0.5$ and training seed $0$, we select backgrounds
at $\sigma_0'^2=0.750$ ($|A|=3$), $0.840$ ($|A|=2$),
$0.930$ ($|A|=1$), and $1.020$ ($|A|=0$).
The active encoder is frozen, the collapsed posteriors are set to
standard normals, and the decoder is projected as in
Eq.~\eqref{eq:numerical-decoder-projection}.
These operations define approximate stationary backgrounds; the
conditional residual mean provides a diagnostic of their stationarity
and finite-sampling error.

Each measurement repeat uses $B=4096$ data points, $L=1024$ active-latent
samples, and $K_{\rm MC}=128$ collapsed-coordinate samples.
We retain the leading centered empirical eigenfunction $\psi$,
normalized by $B^{-1}\sum_b\psi(\bm x_b)^2=1$, and its eigenvalue
$\lambda_{\rm mode}$, and define the decoder partner
$\bm\chi=\widehat{\mathcal R}_A\psi/\sqrt{\lambda_{\rm mode}}$.
Here $\lambda_{\rm mode}$ is the finite-sample spectral estimate paired
with the constructed mode, whereas $\hat\Lambda_*^{(A)}$ in the
continuation analysis denotes the extrapolated estimate.
Power iteration uses two initial vectors and at most $300$ steps per
initialization, with relative eigenvector-residual tolerance $10^{-6}$;
the mode with the larger Rayleigh quotient is retained.

At each probe variance $\sigma'^2$, let $(u,v)$ be the unit lower-eigenvalue
vector of the mean-channel block in Eq.~\eqref{eq:restricted-mass-blocks},
with $\Lambda_\alpha^{(A)}$ replaced by $\lambda_{\rm mode}$.
For one collapsed coordinate $c$, we impose
\begin{equation}
 \mu_c(\bm x;t)=t v\psi(\bm x),\qquad
 \bm f(\bm z;t)=\widehat{\bm f}^{A}(\bm z_A)
                   +t u z_c\bm\chi(\bm z_A),\qquad \sigma_c=1.
 \label{eq:numerical-coupled-perturbation}
\end{equation}
All remaining encoder fields are held fixed. An independent set of
$16384$ joint samples $(\bm x,\bm z_A)$ from the same empirical data
distribution evaluates the negative ELBO, using common samples for
the perturbed and unperturbed fields. Integration over $z_c$ uses its
exact Gaussian first and second moments. The measured curvature is
\begin{equation}
 \kappa(t)=\frac{\mathcal F(t)+\mathcal F(-t)-2\mathcal F(0)}{t^2}.
 \label{eq:numerical-direct-curvature}
\end{equation}
The calculation uses double precision and amplitudes
$t=0.01,0.02,0.04,0.08$. Figure~\ref{fig:direct-curvature} shows the raw
differences at $t=0.01$, including their finite-amplitude correction
of order $t^2$, for $21$ probe ratios
$\sigma'^2/\lambda_{\rm mode}=0.90,0.91,\ldots,1.10$.
Four independently seeded measurement repeats determine means and
standard errors; probe points within a repeat share samples.

The direct curvatures change sign near the predicted spectral edge
for each active dimension, with small offsets near marginality.
This comparison tests the constructed mean-channel direction on a
fixed projected background. The background remains fixed as
$\sigma'^2$ varies, so the scan probes local curvature rather than
tracking a sequence of stationary solutions. Finite data and quadrature
budgets and approximate background stationarity limit the comparison.

\end{document}